\documentclass[10pt,twocolumn,letterpaper]{article}

\usepackage[pagenumbers]{iccv} % To force page numbers, e.g. for an arXiv version

\definecolor{iccvblue}{rgb}{0.21,0.49,0.74}
\usepackage[pagebackref,breaklinks,colorlinks,allcolors=iccvblue]{hyperref}
\usepackage[most]{tcolorbox}
\usepackage{pifont}
\usepackage[utf8]{inputenc}
\usepackage[T1]{fontenc}
\usepackage{bbding}
\usepackage{tikz}
\usepackage{amsfonts}
\usepackage{amssymb}
\usepackage{makecell}
\usepackage{tabularx}
\usepackage{booktabs}
\newcommand{\bcircled}[1]{%
  \tikz[baseline=(char.base)]{%
    \node[shape=circle, fill=black, text=white, inner sep=1pt] (char) {\small\textbf{#1}};%
  }%
}
\newcommand\blfootnote[1]{%
  \begingroup
  \renewcommand\thefootnote{}%
  \footnotetext{#1}%
  \addtocounter{footnote}{-1}%
  \endgroup
}
\def\paperID{*****} % *** Enter the Paper ID here
\def\confName{ICCV}
\def\confYear{2025}

\title{Prompt4Trust: A Reinforcement Learning Prompt Augmentation Framework for Clinically-Aligned Confidence Calibration in Multimodal \\Large Language Models}

\author{Anita Kriz\footnotemark[1] \quad Elizabeth Laura Janes\footnotemark[1] \quad Xing Shen\footnotemark[1]\addtocounter{footnote}{1}\:\:\thanks{Corresponding author.} \quad Tal Arbel \\
Centre for Intelligent Machines, McGill University \quad Mila -- Quebec AI Institute \\
{\tt\small \{anita.kriz,elizabeth.janes,xing.shen\}@mail.mcgill.ca} \quad {\tt\small tal.arbel@mcgill.ca} \\
\footnotemark[1]\;\;Equal contribution
}

\begin{document}
\maketitle

\begin{abstract}
Multimodal large language models (MLLMs) hold considerable promise for applications in healthcare. However, their  deployment in safety-critical settings is hindered by two key limitations: (i) sensitivity to prompt design, and (ii) a tendency to generate incorrect responses with high confidence. As clinicians may rely on a model's stated confidence to gauge the reliability of its predictions, it is especially important that when a model expresses high confidence, it is also highly accurate. We introduce \textbf{Prompt4Trust}, the first reinforcement learning (RL) framework for prompt augmentation targeting confidence calibration in MLLMs. A lightweight LLM is trained to produce context-aware auxiliary prompts that guide a downstream task MLLM to generate responses in which the expressed confidence more accurately reflects predictive accuracy. Unlike conventional calibration techniques, \textbf{Prompt4Trust} specifically prioritizes aspects of calibration most critical for safe and trustworthy clinical decision-making. Beyond improvements driven by this clinically motivated calibration objective, our proposed method also improves task accuracy, achieving state-of-the-art medical visual question answering (VQA) performance on the PMC-VQA benchmark, which is composed of multiple-choice questions spanning diverse medical imaging modalities. Moreover, our framework trained with a small downstream task MLLM showed promising zero-shot generalization to larger MLLMs in our experiments, suggesting the potential for scalable calibration without the associated computational costs. This work demonstrates the potential of automated yet human-aligned prompt engineering for improving the the trustworthiness of MLLMs in safety critical settings. Our codebase can be found at \url{https://github.com/xingbpshen/prompt4trust}.
\end{abstract}

\section{Introduction}
\label{sec:intro}
\blfootnote{© 2025 IEEE. Personal use of this material is permitted. Permission from IEEE must be obtained for all other uses, in any current or future media, including reprinting/republishing this material for advertising or promotional purposes, creating new collective works, for resale or redistribution to servers or lists, or reuse of any copyrighted component of this work in other works.}Multimodal large language models (MLLMs) demonstrate a remarkable ability to reason about complex and diverse medical information~\cite{Singhal2022LargeLM,Thirunavukarasu2023LargeLM,Moor2023FoundationMF}. Unlike traditional AI tools for medical imaging, MLLMs combine analytical abilities with interpretable dialogue, providing intuitive and interactive decision support that will seamlessly integrate with established clinical workflows. Although the integration of MLLMs into real healthcare settings is still in its nascent stages, their potential is increasingly evident: from diagnosing pathological features in radiology scans, to identifying and explaining histopathological findings from microscopy images, MLLMs are poised to revolutionize clinical practice.

Despite their impressive capabilities, MLLMs face two significant challenges to their deployment in safety-critical settings such as healthcare. First, MLLM output quality is highly sensitive to prompt design, with minor semantic changes in prompts leading to substantial response variability. This makes prompt engineering an important but time-consuming and non-trivial step~\cite{lyu2024calibrating, zhao2024fact}. Second, it is well-established that LLMs hallucinate, propagate bias, and produce harmful content~\cite{hallucination_survey,semantic_entropy}, often while presenting their inaccurate outputs as fact. This inherent overconfidence is particularly problematic in clinical contexts, where clinicians may rely on a model's stated confidence to gauge the reliability of its predictions. As such, calibration techniques for clinical settings should be tailored to clinical needs, prioritizing conservative confidence under uncertainty and high confidence in accurate predictions~\cite{nair2020exploring,shen2025improving}, rather than targeting confidence-accuracy alignment across all confidence levels.

Existing confidence calibration techniques attempt to address these challenges through manually designed prompting strategies such as verbalized self-assessments~\cite{xiong2024can}, sample consistency measures~\cite{lyu2024calibrating}, and reflective prompting~\cite{zhao2024fact,tian2023just, huang2024verbalized}. Verbalized confidence methods instruct models to explicitly state their confidence levels, resulting in outputs that align with human expression but still tend to be overconfident~\cite{xiong2024can}.  While Xiong \etal~\cite{xiong2024can} unify these approaches into a comprehensive framework employing chain-of-thought prompting, self-random sampling for consistency evaluation, and response aggregation, this requires computationally expensive sampling. Most critically, all existing confidence calibration methods rely on labor-intensive manual prompt engineering to develop prompts and do not explore how the prompts themselves could be directly \textit{learned} to optimize, not only standard, but clinically informed aspects of calibration.

Automated prompt engineering has emerged as a promising alternative to manual prompt engineering, with reinforcement learning (RL) approaches like RLPrompt~\cite{deng_rlprompt_2022} and PRewrite \cite{kong_prewrite_2024} demonstrating that prompts can be optimized to improve task performance. However, these approaches primarily use \textit{accuracy} as a measure of success~\cite{wu2024prompt}, failing to mitigate overconfidence issues that are particularly detrimental in healthcare settings. This gap in the literature
presents an opportunity to leverage RL for aligning MLLM behavior with calibration quality through well-defined reward functions~\cite{ouyang2022training}.

To deliver an automated yet trustworthy solution, we introduce {\it Prompt4Trust}, the first RL approach to prompt augmentation for MLLM confidence calibration. We train a lightweight \textit{Calibration Guidance Prompt (CGP) Generator} to produce prompts that align a \textit{Downstream Task MLLM's} verbalized confidence with predictive accuracy. Built around a clinically motivated calibration objective, our method penalizes overconfident incorrect answers much more than under-confident but correct ones. As a result, rather than seeking calibration in the traditional sense, our model learns to express caution under uncertainty and confidence when correct, aligned with the standards for safe and trustworthy clinical deployment.

Prompt4Trust improves performance on clinically-motivated calibration objectives, outperforming existing calibration baselines by demonstrating both a higher accuracy when highly confident and conservative behavior when uncertain. Prompt4Trust also achieves state-of-the-art visual question-answering (VQA) accuracy on the PMC-VQA~\cite{zhang_pmc-vqa_2024} benchmark, a challenging medical multiple-choice VQA dataset with diverse imaging modalities, using significantly fewer language model parameters (13B in prior work vs. 3.5B, composed of a 1.5B \textit{CGP Generator} and a 2B \textit{Downstream Task MLLM's}). Our calibration strategy transfers effectively to much larger \textit{Downstream Task MLLMs} without additional fine-tuning, enabling scalable calibration improvements. Our results demonstrate the critical importance of our method in medical imaging, where calibration-focused prompt optimization is both unexplored and critically needed. Crucially, Prompt4Trust ensures that high confidence predictions correspond to high accuracy, meeting the standard for clinical deployment.

\section{Problem Formulation: Prompt Augmentation to Improve Calibration} \label{sec:prompt-aug} 

Consider the problem of answering multiple choice questions based on medical images, where the MLLM is explicitly tasked with producing both a prediction and a corresponding confidence score. Confidence calibration, which we define as the alignment of expressed confidence and accuracy, remains a significant challenge for current models. Our goal is to improve calibration via prompt augmentation. In this section, we formalize prompt augmentation in the context of MLLMs, enabling us to motivate our RL–based approach.

\subsection{Formalism} Let a tuple $(v, t \oplus \mathcal{A})$ denote a multimodal input query, where $v$ is the visual input, $t$ is the question in text,  and $\mathcal{A} = \{a_1, a_2, ..., a_k\}$ is an \textit{optional} set of answer choices,where $\oplus$ denotes text concatenation. The standard approach of querying a Downstream Task MLLM $f_\tau$ involves directly inputting $(v, t \oplus \mathcal{A})$ and asking for the answer and its numerical confidence, resulting in a predicted answer $\hat{y}$, and an associated confidence score $\hat{p}$: $(\hat{y}, \hat{p}) = f_\tau(v, t \oplus \mathcal{A})$. This input can be optionally augmented with an auxiliary prompt $c$, resulting in the modified input $(\hat{y}, \hat{p}) = f_\tau(v, t \oplus \mathcal{A} \oplus c)$.

\subsection{Motivation for Prompt Augmentation} To illustrate why an auxiliary prompt $c$ can improve confidence calibration, we highlight two key properties: (i) \textbf{Non-invariance.} Because MLLMs are autoregressive, it follows that $f_\tau(v, t \oplus \mathcal{A}) \neq f_\tau(v, t \oplus \mathcal{A} \oplus c)$. Thus, $c$ can influence both the MLLM's prediction and its confidence. (ii) \textbf{Orderability.} Prior work~\cite{zhao2024fact} shows that different prompts yield different calibration behavior. For any $c_1, c_2 \in \mathcal{C}$, where $\mathcal{C}$ is the space of candidate prompts, we have $u(f_\tau(v, t \oplus \mathcal{A} \oplus c_1)) \neq u(f_\tau(v, t \oplus \mathcal{A} \oplus c_2))$, where $u(\cdot)$ measures instance-level calibration. This implies that some prompts are more effective than others, making it meaningful to find prompts that improve calibration.

\subsection{The Limitation of Fixed Prompt Augmentation} A straightforward strategy to approach confidence calibration is to append a \textit{fixed} auxiliary prompt to the input. For example, one might define $c=$ \texttt{"think step-by-step, do not be over-confident"}, with the expectation that the Downstream Task MLLM $f_\tau$ will follow this instruction and produce conservative confidence estimates. However, this heuristic approach suffers from two key limitations: (i) The input text $t \oplus \mathcal{A}$ and fixed auxiliary prompt $c$ are independent; consequently, the auxiliary prompt does not leverage the specific content of $t \oplus \mathcal{A}$. (ii) A fixed prompt often cannot generalize well across distributions of input text $t \oplus \mathcal{A}$~\cite{lyu2024calibrating, zhao2024fact}.

\section{Prompt4Trust}
\label{sec:method}

\begin{figure*}[!t]
    \centering
    \includegraphics[width=.9\textwidth]{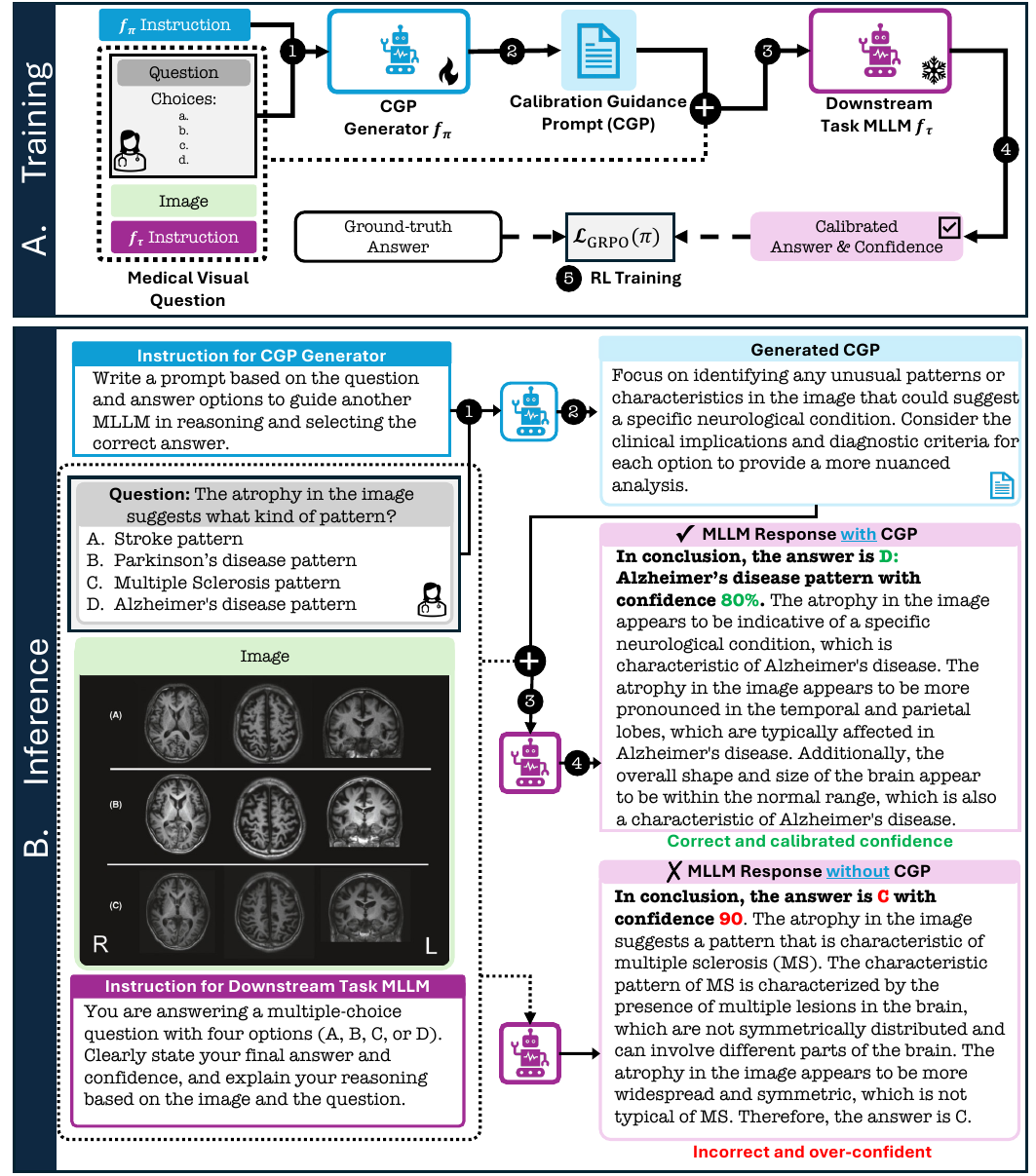}
    \caption{\textbf{Overview of the Prompt4Trust Framework for Medical Visual Question Answering.} 
    \textbf{(A)} \textit{Training pipeline.} \protect\bcircled{1}~The Calibration Guidance Prompt (CGP) Generator receives the textual elements of a medical visual question, namely the question, multiple-choice options, and an instruction. \protect\bcircled{2}~ The CGP Generator subsequently produces the CGP. \protect\bcircled{3}~The CGP is then appended to the question, multiple choice options, and the downstream task instruction, along with the associated medical image, and passed to the \textit{Downstream Task MLLM}. \protect\bcircled{4}~The \textit{Downstream Task MLLM} produces an answer and confidence score, which are compared to the ground-truth answer to compute a reward. 
    The \textit{Downstream Task MLLM} reports its confidence as a score out of 100. $\hat{p}$ is defined by converting this score to a decimal.
    \protect\bcircled{5}~This reward is used to optimize the CGP Generator via the GRPO reinforcement learning objective.
    \textbf{(B)} \textit{Inference on a Sample from the PMC-VQA dataset} \cite{zhang_pmc-vqa_2024}. At inference time, Prompt4Trust follows steps \protect\bcircled{1}–\protect\bcircled{4}, resulting in the \Checkmark~MLLM Response with CGP, which produces the \textbf{\textcolor{ForestGreen}{correct answer and calibrated confidence}}. The Generated CGP text was abbreviated for illustration purposes. For comparison, we show the \ding{55}~MLLM Response without CGP yields an \textbf{\textcolor{red}{incorrect and overconfident}} response, illustrating the effectiveness of the Prompt4Trust framework.}
    \label{fig:method}
\end{figure*}

We propose Prompt4Trust, a framework for \textit{learning} context-aware auxiliary prompts tailored to each input query, which we call \textit{Calibration Guidance Prompts} (CGPs), and present our RL-based framework for doing so.

\subsection{Learning to Generate Context-Aware CGPs} Prompt4Trust is centered around a lightweight \textit{Prompt Generator} $f_\pi$ that takes the textual components of the multimodal query as input to produce a CGP: $c = f_\pi(t \oplus \mathcal{A})$. This enables the generation of the CGP, $c$, to be adapted to each individual query before being passed to the frozen \textit{Downstream Task MLLM}  $f_\tau$ which returns both the predicted answer and its confidence: $(\hat{y}, \hat{p})= f_\tau(v,t \oplus \mathcal{A} \oplus c)$. An overview of the training architecture for this framework is shown in Figure~\ref{fig:method} A.

\subsection{Training CGP Generator Using RL} 
Given a visual-textual input pair $(v,t \oplus \mathcal{A})$ with ground-truth answer $y\in\mathcal{A}$, we generate a CGP $c=f_\pi(t \oplus \mathcal{A})$ and obtain the downstream model's prediction $(\hat{y}, \hat{p}) = f_\tau(v, t \oplus \mathcal{A} \oplus c)$, where $\hat{y}$ is the predicted answer and $\hat{p}$ is the verbalized confidence. Our reward function is defined in Equation \ref{eq:reward}, where $r$ is a function of $y$, $\hat{y}$, and $\hat{p}$:
\begin{equation}
r =
\begin{cases}
\log\big(\min\{1, \max\{\hat{p}, \epsilon\} \} \big), & \text{if } \hat{y} = y \\
\log\big(\min\{1, \max\{1 \!-\! \hat{p}, \epsilon\} \} \big) \! - \! 1, & \text{if } \hat{y} \neq y \\
\log(\varepsilon_\text{penalty}), & \text{else}
\end{cases}
\label{eq:reward}
\end{equation}
This reward function encourages $f_\pi$ to generate CGPs that guide $f_\tau$ toward well-calibrated confidence, while prioritizing the behavior that is most critical to clinical decision-making: when a model expresses high confidence, it should be correct. Correct predictions are rewarded in proportion to their confidence, with higher $\hat{p}$ values yielding higher rewards. Conversely, incorrect predictions are penalized more as confidence increases (via the $1-\hat{p}$ term), and an additional $-1$ penalty is applied. This asymmetry in the reward function differentiates our approach from traditional definitions of calibration and serves two key purposes: (i) it discourages overconfidence errors by  assigning the largest penalties to overconfidently incorrect answers, and (ii) it encourages confident expression when the model is correct by reserving the maximal reward for highly confident correct predictions. While some correct predictions may be expressed with lower confidence due to this reward scheme, this behavior is desirable in high stakes settings where conservative uncertainty is preferred over unjustified confidence.

Responses are considered invalid if both answer and confidence are not reported, or if they are not reported in a format that we can parse based on our given instructions. We introduce $\epsilon=1\times10^{-10}$ for numerical stability and $\varepsilon_\text{penalty}=1\times10^{-12}$ which heavily penalizes invalidly formatted outputs with $\log(\varepsilon_\text{penalty})$. 

\subsection{Learning the Optimal Policy} We adopt the standard Group-Relative Policy Optimization (GRPO)~\cite{shao2024deepseekmath} to learn the parameters $\pi$ of the policy model $f_\pi$.  GRPO provides computational efficiency by avoiding the need for a separate value function, unlike PPO, while maintaining stable training dynamics. The training objective is $\mathcal{L}(\pi) := \mathcal{L}_\mathrm{GRPO}(\pi)$.

\section{Experiments}
\begin{table*}[t]
\centering
\caption{
Expected Calibration Error (ECE), Brier score, and accuracy evaluated on the \textbf{test set}. Best scores are shown in \textbf{bold}. The previous state-of-the-art (SOTA) accuracy on the \textit{PMC-VQA-test} set is 42.3\% (MedVInT with a 13B language model \cite{zhang_pmc-vqa_2024,wu2024pmc}), as reported on the PMC-VQA leaderboard~\cite{zhang_pmc-vqa_2024}; \textbf{Prompt4Trust achieves a new SOTA with significantly fewer language model parameters (3.5B in total)}.\\ \small* ECE and Brier score are computed over valid samples only; accuracy includes all 2,000 test samples, treating invalid outputs as incorrect.
}
\begin{tabular*}{.8\textwidth}{@{\extracolsep{\fill}}lcccc}
\toprule
\textbf{Method} & \textbf{ECE\textsuperscript{*} $\downarrow$} & \textbf{Brier\textsuperscript{*} $\downarrow$}  & \textbf{Valid Samples\textsuperscript{*}} & \textbf{Accuracy (\%) $\uparrow$} \\
\midrule
Verbalized & 0.517 & 0.523 & 1803 & 40.40 \\
Verbalized + Fixed-PA & 0.493 & 0.496 & 1923 & 44.25 \\
Consistency & 0.265 & 0.323 & 1866 & 36.60 \\
Avg-Conf & 0.265 & 0.322& 1866 & 36.60  \\
Prompt4Trust (Ours) & \textbf{0.163} & \textbf{0.264} & \textbf{1971} & \textbf{45.95} \\
\bottomrule
\end{tabular*}
\label{tab:final_test_results}
\end{table*}
In clinical settings, decision-making will prioritize the responses that MLLMs express with high confidence, making the high-confidence region most informative of whether expressed confidence can be trusted to guide clinical decisions. Conversely, in low accuracy regions, it is crucial that a MLLM avoids exhibiting overconfidence, providing an essential safety margin in clinical deployment.
Based on this motivation, we design our experiments to investigate two key questions:
\begin{enumerate}
    \item \textbf{Benchmark Experiment:} Can this clinically motivated calibration objective for a \textit{Downstream Task MLLM} be improved by using context-aware CGPs from our \textit{CGP Generator} compared to established baseline methods? (Section~\ref{subsec:main})
    \item \textbf{Generalizability Experiment:} Can our trained \textit{CGP Generator} produce CGPs that generalize to unseen, larger \textit{Downstream Task MLLMs} with different architectures? (Section~\ref{subsec:generalize})
\end{enumerate}

\subsection{Experimental Setup}
\subsubsection{Models}
Our framework utilizes four distinct models across the experiments. For the \textit{CGP Generator}, we fine-tune instruction-tuned Qwen2.5-1.5B-Instruct~\cite{qwen2025qwen25technicalreport, qwen2.5, qwen2} to produce context-aware CGPs. In the benchmark experiment, we use instruction-tuned Qwen2-VL-2B-Instruct~\cite{wang2024qwen2vlenhancingvisionlanguagemodels, Qwen-VL} as the \textit{Downstream Task MLLM} for visual question answering. To evaluate the generalizability of Prompt4Trust across different architectures, we employ Qwen2.5-VL-7B-Instruct~\cite{wang2024qwen2vlenhancingvisionlanguagemodels, Qwen-VL, qwen2.5-VL} and InternVL3-14B-Instruct~\cite{chen2024expanding, wang2024mpo, chen2024far, chen2024internvl} as alternative \textit{Downstream Task MLLMs}\footnote{All models and data are used in accordance with their license agreements and ethical guidelines.}.

\subsubsection{Dataset} 
% \noindent\textbf{Dataset.}
We utilize PMC-VQA~\cite{zhang_pmc-vqa_2024, zhang_development_2024}, a diverse multiple-choice medical visual question answering (VQA) dataset consisting of 227,000 multiple-choice questions, each paired with a 2D medical image, four possible answers, and a ground truth label. The dataset includes images selected from academic papers and are 80\% radiological images, including X-rays, MRIs, CT scans, and PET scans, as well as pathology, microscopy, and signals images~\cite{zhang_pmc-vqa_2024, zhang_development_2024}. Compared to previous benchmarks, PMC-VQA is very challenging: state-of-the-art models perform only slightly better than random~\cite{zhang_pmc-vqa_2024, zhang_development_2024}. 

\noindent\textbf{Data splits.}
We train our model with a randomly selected subset of 5{,}000 questions from the PMC-VQA training split. Hyperparameter tuning is conducted on a validation set of 1{,}000 randomly selected samples. Our final results are evaluated on the \textit{PMC-VQA-test} set which was curated and manually reviewed by Zhang \etal~\cite{zhang_pmc-vqa_2024, zhang_development_2024}.

\noindent\textbf{Framework adaptability. } Developed on PMC-VQA’s 2D images and multiple-choice format, our framework extends to any visual modality supported by the \textit{Downstream Task MLLM}, provided it accepts text prompts and ground-truth labels are available for training.

\subsubsection{Hyperparameters and Training}
During training, the \textit{CGP Generator} is finetuned while all \textit{Downstream Task MLLMs} remain frozen. We adopt the default GRPO configuration~\cite{shao2024deepseekmath}. During training, we sample eight candidate completions per prompt to enable reward computation under GRPO. We use the AdamW optimizer~\cite{loshchilovdecoupled} with a learning rate of $1\times10^{-6}$ and apply no reward normalization, following~\cite{liu2025understandingr1zeroliketrainingcritical}. 
The prompt and completion lengths are set to a maximum of 512 and 256 tokens, respectively. Training is performed with four iterations per batch and a per-device batch size of 8.

A grid search over \texttt{temperature}, \texttt{top-k}, and \texttt{beta} is conducted to obtain optimal hyperparameter configuration. We report all results using the optimal hyperparameter configuration ($\texttt{beta} = 0.04, \texttt{top-k} = 100, \texttt{temperature} =0.6$) which favorably balances calibration and accuracy.

\subsubsection{Baselines}
We evaluate our method and compare it against several baselines:
\begin{itemize}
    \item \textbf{Verbalized:} directly queries the \textit{Downstream Task MLLM} to output its answer and verbalize its confidence.
    \item \textbf{Verbalized + Fixed-PA:} extends the above approach by using the fixed auxiliary prompt augmentation $c=$ \texttt{"think step-by-step, do not be over-confident"}.
    \item \textbf{Consistency:} measures confidence by evaluating agreement among multiple model responses, calculating confidence as the proportion of sampled responses that match the original answer. It is applied with zero-shot chain-of-thought prompting and self-random sampling, with each query sampled 21 times~\cite{xiong2024can}.
    \item \textbf{Average Confidence (Avg-Conf):} combines response consistency with verbalized confidence scores as weighting factors across sampled responses. It uses the same zero-shot chain-of-thought prompting and sampling strategy as Consistency.
\end{itemize}

\subsubsection{Evaluation Metrics}
We assess confidence calibration using the following metrics: (i) \textbf{Expected Calibration Error (ECE)}~\cite{pakdaman_naeini_obtaining_2015}, which measures the average discrepancy between verbalized confidence and empirical accuracy; and (ii) \textbf{Brier score}~\cite{brier_verification_1950}, which evaluates the mean squared difference between the predicted probability (defined here as the verbalized confidence) and the actual outcome. ECE and Brier scores can range from zero to one, with scores of zero indicating optimal calibration. Additionally, we report \textbf{accuracy}, defined as the proportion of instances in which the downstream model selects the correct answer from the set of multiple-choice options, to assess overall task performance.

\noindent\textbf{Metric Reporting Note.} Due to Qwen2-VL-2B-Instruct~\cite{wang2024qwen2vlenhancingvisionlanguagemodels, Qwen-VL} requiring images to be at least $28\times28$ pixels, three samples from \textit{PMC-VQA-test} were excluded. Therefore, our final results for ECE and Brier score are reported on 1{,}997 out of the 2{,}000 original samples in \textit{PMC-VQA-test}. To ensure a fair comparison with the PMC-VQA leaderboard, which reports accuracy on the entire set of 2,000 samples, we explicitly treat methods Prompt4Trust, Verbalized, Consistency, and Avg-Conf as having made \textit{incorrect} predictions on those three excluded samples.

If the \textit{Downstream Task MLLM} outputs are reported without the predicted answer or confidence estimate, or if these are output in a manner that cannot be parsed, we consider the response to be invalid. In the benchmark experiment, shown in Table \ref{tab:final_test_results}, we count invalid responses as incorrect when computing accuracy, however, invalid samples are omitted from the ECE and Brier score calculations.

\begin{table}[t]
\centering
\caption{Average confidence on incorrectly answered questions. \textbf{Prompt4Trust} exhibits a desirable property by assigning \textcolor{ForestGreen}{low confidence with low standard deviation to incorrect predictions}. In contrast, \textbf{Verbalized} and \textbf{Verbalized + Fixed-PA} assign \textcolor{red}{high confidence despite being wrong}, \textbf{Consistency} and \textbf{Avg-Conf} assign confidence with high standard deviation when being wrong.}
\begin{tabular*}{0.9\linewidth}{@{\extracolsep{\fill}}lc@{}}
\toprule
\textbf{Method} & \makecell{\textbf{Avg. Confidence on} \\ \textbf{Incorrect Answers} {\scriptsize $\pm$ std}} \\ \midrule
Verbalized & \textcolor{red}{0.960} {\scriptsize $\pm 0.135$} \\
Verbalized + Fixed-PA & \textcolor{red}{0.940} {\scriptsize $\pm 0.155$} \\
Consistency & 0.556 \textcolor{red}{\scriptsize $\pm 0.301$} \\
Avg-Conf & 0.555 \textcolor{red}{\scriptsize $\pm 0.301$} \\
Prompt4Trust (Ours) & \textcolor{ForestGreen}{0.571} \textcolor{ForestGreen}{\scriptsize $\pm 0.180$} \\ \bottomrule
\end{tabular*}
\label{tab:avg-conf-incorrect}
\end{table}

\begin{table}[]
\centering
\caption{Accuracy of high-confidence ($\text{confidence}\geq 0.85$) predictions for each method. \textbf{Prompt4Trust} outperforms all baseline methods when the predictions are made with high confidence.}
\label{tab:acc-on-high-conf}
\begin{tabular}{lc}
\toprule
\textbf{Method} & \begin{tabular}[c]{@{}c@{}}\textbf{Accuracy (\%) $\uparrow$}\\(on conf. $\geq 0.85$)\end{tabular} \\ \midrule
Verbalized            & 44.74 \\
Verbalized + Fixed-PA & 46.99 \\
Consistency           & 50.71 \\
Avg-Conf              & 50.49 \\
Prompt4Trust (Ours)   & \textbf{76.87} \\ \bottomrule
\end{tabular}
\end{table}

\begin{table*}[h]
\centering
\caption{Zero-shot generalization of the trained \textit{CGP Generator} to larger \textit{Downstream Task MLLMs} (Qwen2.5-VL-7B-Instruct~\cite{wang2024qwen2vlenhancingvisionlanguagemodels, Qwen-VL, qwen2.5-VL} and InternVL3-14B-Instruct~\cite{chen2024expanding, wang2024mpo, chen2024far, chen2024internvl}). The \textit{CGP Generator} was originally trained with Qwen2-VL-2B-Instruct~\cite{wang2024qwen2vlenhancingvisionlanguagemodels, Qwen-VL} and is applied without fine-tuning. Results are shown for Expected Calibration Error (ECE), Brier score, and accuracy. Best results are in \textbf{bold} and second best results are in \underline{underlined}. Results demonstrate improvements in confidence calibration when prompted with CGPs in a zero-shot manner, while maintaining comparable accuracy.\\ \small* ECE and Brier score are computed over valid samples only; accuracy includes all 2,000 test samples, treating invalid outputs as incorrect.}
\begin{tabularx}{0.893\textwidth}{lccccc}
\toprule 
\begin{tabular}[c]{@{}l@{}}\textbf{Downstream MLLM}\\ \textbf{and Method}\end{tabular} & \textbf{ECE\textsuperscript{*} $\downarrow$} & \textbf{Brier\textsuperscript{*} $\downarrow$} & \textbf{Valid Samples\textsuperscript{*}} & \begin{tabular}[c]{@{}c@{}}\textbf{Accuracy (\%) $\uparrow$}\\(on 2,000)\end{tabular} & \begin{tabular}[c]{@{}c@{}}\textbf{Accuracy (\%) $\uparrow$}\\(on conf. $\geq 0.85$)\end{tabular} \\
\midrule
\rowcolor{gray!15}
Qwen2.5-VL-7B-Instruct & & & & & \\
Verbalized & 0.387 & 0.396 & \underline{1993} & \underline{52.15} & 53.05 \\
Verbalized + Fixed-PA & \underline{0.341} & \underline{0.356} & \textbf{1996} & \textbf{52.75} & \underline{55.78} \\
Prompt4Trust (Ours) & \textbf{0.289} & \textbf{0.317} & \textbf{1996} & 48.65 & \textbf{74.65} \\
\midrule
\rowcolor{gray!15}
InternVL3-14B-Instruct  & & & & & \\
Verbalized & 0.371 & 0.377 & \textbf{1997} & \textbf{54.35} & 54.55 \\
Verbalized + Fixed-PA & \underline{0.354} & \underline{0.367} & \textbf{1997} & \textbf{54.35} & \underline{55.17} \\
Prompt4Trust (Ours) & \textbf{0.302} &  \textbf{0.327} & \underline{1995} & \underline{51.17} & \textbf{63.97} \\
\bottomrule
\end{tabularx}
\label{tab:generalizability_results}
\end{table*}

\subsection{Benchmark Experiment: Prompt4Trust Calibration Performance}
\label{subsec:main}
The primary objective of this experiment is to evaluate whether Prompt4Trust improves confidence calibration compared to established baselines while maintaining or improving task accuracy. In particular, we assess whether context-aware CGPs have advantages over fixed prompt augmentations.
\paragraph{Results.}
Table~\ref{tab:final_test_results} reports performance on the held-out test set. The \textbf{Verbalized} baseline exhibits substantially higher ECE and Brier scores, consistent with prior findings that zero-shot prompting for verbalized confidence is insufficient for effective calibration~\cite{xiong2024can}. While \textbf{Verbalized + Fixed-PA} improves calibration relative to Verbalized by adding prompt augmentation, its performance is limited by the augmentation's lack of input-specific context.

The \textbf{Consistency} and \textbf{Avg-Conf} baselines achieve lower calibration error by aggregating over multiple samples, but at the cost of reduced accuracy. In contrast, \textbf{Prompt4Trust} outperforms all baselines in both calibration and accuracy, achieving state-of-the-art results on the \textit{PMC-VQA-test} set according to the leaderboard\footnote{Leaderboard available via the link in~\cite{zhang_pmc-vqa_2024}.} from Zhang \textit{et al.}~\cite{zhang_pmc-vqa_2024}.

\begin{figure}[t]
    \centering
    \includegraphics[width=0.8\linewidth]{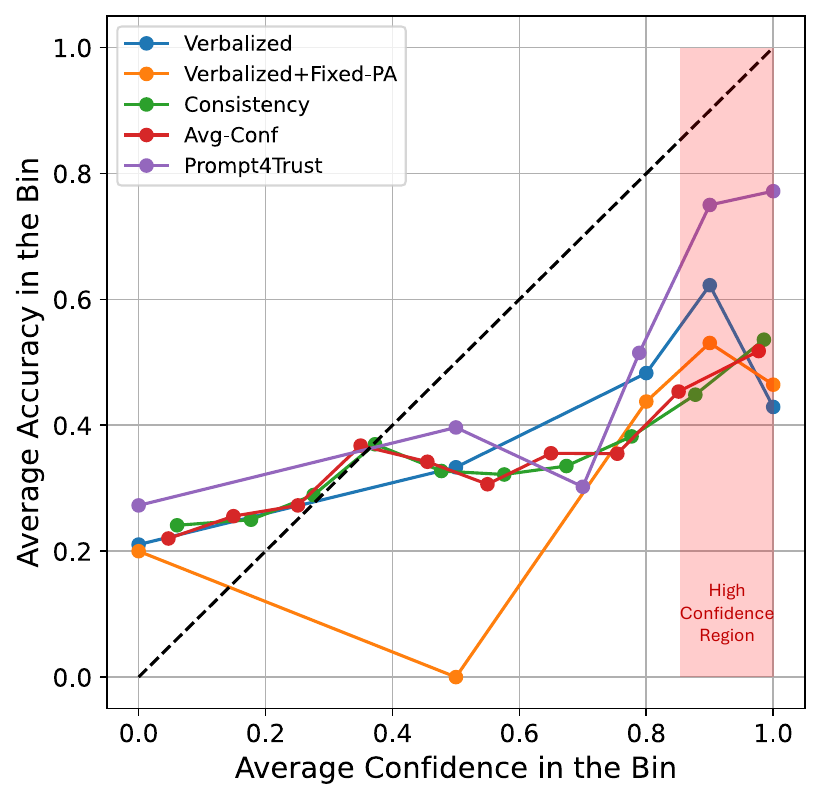}
    \caption{Calibration curve illustrating the relationship between confidence and average accuracy. Perfect calibration is shown by the dashed line. \textbf{Prompt4Trust demonstrates better calibration, particularly in high-confidence region} (e.g., $\text{confidence}\geq 0.85$) \textbf{where trust is most critical for supporting medical decision-making in medical imaging tasks.}}
    \label{fig:calibration-curve}
\end{figure}

The calibration curves of Prompt4Trust and other methods are shown in Figure \ref{fig:calibration-curve}. Prompt4Trust is particularly effective in the region of high confidence predictions: when the \textit{Downstream Task MLLM} expresses high confidence, Prompt4Trust maintains accuracy levels that align with its expressed confidence, outperforming all baselines. In the low confidence region (confidence $\leq 0.4$), Prompt4Trust tends to report confidence levels lower than its actual accuracy, thereby providing a safety margin for medical decision-making. 

Table \ref{tab:avg-conf-incorrect} reports the average confidence expressed in incorrectly answered responses for Prompt4Trust and all baselines. When incorrect, Prompt4Trust exhibits a low confidence ($0.571$) with low standard deviation ($\pm 0.180$). In contrast, even when incorrect, the Verbalized and Verbalized + FixedPA baselines express very high confidence ($0.960$ and $0.940$, respectively), which lacks trustworthiness in safety-critical decisions. Furthermore, the low standard deviation in their confidence estimates indicates a consistent pattern of overconfidence. While Consistency and Avg-Conf produce lower average confidence on incorrect predictions ($0.556$ and $0.555$, respectively), their high standard deviations ($\pm0.301$) indicate inconsistent and unreliable uncertainty estimates.

Table \ref{tab:acc-on-high-conf} reports the average accuracy on samples where the model's expressed confidence is at least $0.85$. Prompt4Trust achieves higher average accuracy compared to all other baselines. This indicates that when Prompt4Trust expresses high confidence, its predictions are more accurate than all other baselines when they express similar levels of high confidence.

Importantly, Prompt4Trust is also inference-efficient: unlike sampling-based approaches, it produces calibrated verbalized confidence estimates with a single forward pass. Figure~\ref{fig:method}B presents qualitative inference results from Prompt4Trust and the Verbalized baseline on a representative sample from the PMC-VQA dataset.

\begin{figure}[h]
    \centering
    \includegraphics[width=0.8\linewidth]{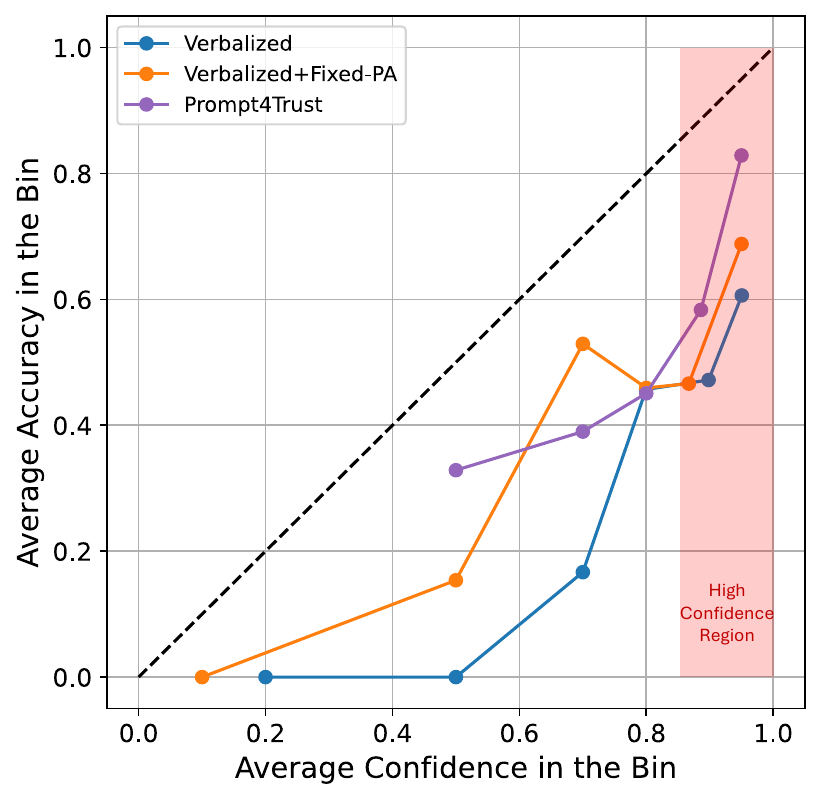}
    \caption{The calibration curve for Qwen2.5-VL-7B-Instruct~\cite{bai2025qwen25vl} as the \textit{Downstream Task MLLM} in the generalizability experiment. \textbf{Prompt4Trust demonstrates better calibration, particularly in high-confidence region} (e.g., $\text{confidence}\geq 0.85$) \textbf{where trust is most critical for supporting medical decision-making.}}
    \label{fig:calibration-curve-qwen25}
\end{figure}

\subsection{Generalizability Experiment: Prompt4Trust Cross-Model Performance}
\label{subsec:generalize}
This experiment evaluates the cross-architecture generalization of our trained \textit{CGP Generator} on a substantially larger \textit{Downstream Task MLLM} without additional fine-tuning. This tests a practical training framework that could avoid the computational bottleneck of directly training with large computationally intensive architectures.

\paragraph{Results.}
Table \ref{tab:generalizability_results} demonstrates the zero-shot generalizability of Prompt4Trust to both Qwen2.5-VL-7B-Instruct~\cite{wang2024qwen2vlenhancingvisionlanguagemodels, Qwen-VL, qwen2.5-VL} and InternVL3-14B-Instruct~\cite{chen2024expanding, wang2024mpo, chen2024far, chen2024internvl}. Prompt4Trust achieves reduced ECE and Brier score compared to all baselines, with comparable accuracy. This result highlights the practical value of our framework: the \textit{CGP Generator} can be trained with smaller, accessible \textit{Downstream Task MLLMs}, and then applied to more powerful models that are too large to involve directly in training. Moreover, in the high-confidence region (confidence $\geq 0.85$), Prompt4Trust improves accuracy across both \textit{Downstream Task MLLMs} relative to all baselines. 
The calibration curve in Figure \ref{fig:calibration-curve-qwen25} illustrates this effect on Qwen2.5-VL-7B-Instruct~\cite{wang2024qwen2vlenhancingvisionlanguagemodels, Qwen-VL, qwen2.5-VL}, further highlighting Prompt4Trust's effectiveness in ensuring accurate responses when the \textit{Downstream Task MLLM} expresses highly confident.

Although InternVL3-14B-Instruct~\cite{chen2024expanding, wang2024mpo, chen2024far, chen2024internvl} exhibits smaller calibration performance gains with Prompt4Trust compared to Qwen2.5-VL-7B-Instruct~\cite{wang2024qwen2vlenhancingvisionlanguagemodels, Qwen-VL, qwen2.5-VL}, it highlights the potential for Prompt4Trust to generalize to a large model with a different architecture than the \textit{Downstream Task MLLM} used during training, and motivates future work to further improve Prompt4Trust's zero-shot generalizability.

\section{Conclusion and Future Work}
In this work we introduce Prompt4Trust, the first RL-based prompt augmentation framework targeting clinically aligned confidence calibration in MLLMs applied to medical VQA. We demonstrate that by learning context-aware CGPs, our method improves both calibration and accuracy, with calibration benefits transferring to unseen, larger \textit{Downstream Task MLLMs}. Our results demonstrate the importance of developing calibration methods that align with the domain-specific needs in the healthcare setting, where reliable high confidence predictions and conservative low-confidence behavior is more valuable than marginal accuracy gains or standard calibration metrics. Future directions include extending Prompt4Trust into agentic frameworks, where increased interaction could further enhance calibration quality.

\section*{Acknowledgments}
This work was supported in part by the Natural Sciences and Engineering Research Council of Canada (NSERC), in part by the Canadian Institute for Advanced Research (CIFAR) Artificial Intelligence Chairs Program, in part by the Mila -- Quebec Artificial Intelligence Institute, in part by the compute resources provided by Mila (mila.quebec), in part by the Mila-Google Research Grant, in part by the Fonds de recherche du Québec, in part by the Canada First Research Excellence Fund, awarded to the Healthy Brains, Healthy Lives initiative at McGill University, and in part by the Department of Electrical and Computer Engineering at McGill University.

We would also like to acknowledge the Reinforcement Learning course at McGill University, taught by Doina Precup and Isabeau Prémont-Schwarz, for the valuable insights and motivations that it provided.

{
    \small
    \bibliographystyle{ieeenat_fullname}
    \bibliography{egbib}

\begin{thebibliography}{35}
\providecommand{\natexlab}[1]{#1}
\providecommand{\url}[1]{\texttt{#1}}
\expandafter\ifx\csname urlstyle\endcsname\relax
  \providecommand{\doi}[1]{doi: #1}\else
  \providecommand{\doi}{doi: \begingroup \urlstyle{rm}\Url}\fi

\bibitem[Bai et~al.(2023)Bai, Bai, Yang, Wang, Tan, Wang, Lin, Zhou, and Zhou]{Qwen-VL}
Jinze Bai, Shuai Bai, Shusheng Yang, Shijie Wang, Sinan Tan, Peng Wang, Junyang Lin, Chang Zhou, and Jingren Zhou.
\newblock Qwen-vl: A versatile vision-language model for understanding, localization, text reading, and beyond.
\newblock \emph{arXiv preprint arXiv:2308.12966}, 2023.

\bibitem[Bai et~al.(2025)Bai, Chen, Liu, Wang, Ge, Song, Dang, Wang, Wang, Tang, et~al.]{bai2025qwen25vl}
Shuai Bai, Keqin Chen, Xuejing Liu, Jialin Wang, Wenbin Ge, Sibo Song, Kai Dang, Peng Wang, Shijie Wang, Jun Tang, et~al.
\newblock Qwen2.5-vl technical report.
\newblock \emph{arXiv preprint arXiv:2502.13923}, 2025.

\bibitem[Brier(1950)]{brier_verification_1950}
Glenn~W. Brier.
\newblock {Verification} {of} {Forecasts} {Expressed} {in} {Terms} {of} {Probability}.
\newblock \emph{Monthly Weather Review}, 78\penalty0 (1), 1950.

\bibitem[Chen et~al.(2024{\natexlab{a}})Chen, Wang, Cao, Liu, Gao, Cui, Zhu, Ye, Tian, Liu, et~al.]{chen2024expanding}
Zhe Chen, Weiyun Wang, Yue Cao, Yangzhou Liu, Zhangwei Gao, Erfei Cui, Jinguo Zhu, Shenglong Ye, Hao Tian, Zhaoyang Liu, et~al.
\newblock Expanding performance boundaries of open-source multimodal models with model, data, and test-time scaling.
\newblock \emph{arXiv preprint arXiv:2412.05271}, 2024{\natexlab{a}}.

\bibitem[Chen et~al.(2024{\natexlab{b}})Chen, Wang, Tian, Ye, Gao, Cui, Tong, Hu, Luo, Ma, et~al.]{chen2024far}
Zhe Chen, Weiyun Wang, Hao Tian, Shenglong Ye, Zhangwei Gao, Erfei Cui, Wenwen Tong, Kongzhi Hu, Jiapeng Luo, Zheng Ma, et~al.
\newblock How far are we to gpt-4v? closing the gap to commercial multimodal models with open-source suites.
\newblock \emph{arXiv preprint arXiv:2404.16821}, 2024{\natexlab{b}}.

\bibitem[Chen et~al.(2024{\natexlab{c}})Chen, Wu, Wang, Su, Chen, Xing, Zhong, Zhang, Zhu, Lu, et~al.]{chen2024internvl}
Zhe Chen, Jiannan Wu, Wenhai Wang, Weijie Su, Guo Chen, Sen Xing, Muyan Zhong, Qinglong Zhang, Xizhou Zhu, Lewei Lu, et~al.
\newblock Internvl: Scaling up vision foundation models and aligning for generic visual-linguistic tasks.
\newblock In \emph{Proceedings of the IEEE/CVF Conference on Computer Vision and Pattern Recognition}, pages 24185--24198, 2024{\natexlab{c}}.

\bibitem[Deng et~al.(2022)Deng, Wang, Hsieh, Wang, Guo, Shu, Song, Xing, and Hu]{deng_rlprompt_2022}
Mingkai Deng, Jianyu Wang, Cheng-Ping Hsieh, Yihan Wang, Han Guo, Tianmin Shu, Meng Song, Eric Xing, and Zhiting Hu.
\newblock {RLPrompt}: {Optimizing} {Discrete} {Text} {Prompts} with {Reinforcement} {Learning}.
\newblock In \emph{Proceedings of the 2022 {Conference} on {Empirical} {Methods} in {Natural} {Language} {Processing}}, pages 3369--3391, Abu Dhabi, United Arab Emirates, 2022. Association for Computational Linguistics.

\bibitem[Farquhar et~al.(2024)Farquhar, Kossen, Kuhn, and Gal]{semantic_entropy}
Sebastian Farquhar, Jannik Kossen, Lorenz Kuhn, and Yarin Gal.
\newblock Detecting hallucinations in large language models using semantic entropy.
\newblock \emph{Nature}, 630\penalty0 (8017):\penalty0 625--630, 2024.

\bibitem[Huang et~al.(2024)Huang, Shen, Wang, Meng, Liu, Wang, and Bhatt]{huang2024verbalized}
Hengguan Huang, Xing Shen, Songtao Wang, Lingfa Meng, Dianbo Liu, Hao Wang, and Samir Bhatt.
\newblock Verbalized probabilistic graphical modeling.
\newblock \emph{arXiv preprint arXiv:2406.05516}, 2024.

\bibitem[Huang et~al.(2025)Huang, Yu, Ma, Zhong, Feng, Wang, Chen, Peng, Feng, Qin, and Liu]{hallucination_survey}
Lei Huang, Weijiang Yu, Weitao Ma, Weihong Zhong, Zhangyin Feng, Haotian Wang, Qianglong Chen, Weihua Peng, Xiaocheng Feng, Bing Qin, and Ting Liu.
\newblock A survey on hallucination in large language models: Principles, taxonomy, challenges, and open questions.
\newblock \emph{ACM Transactions on Information Systems}, 43\penalty0 (2):\penalty0 1–55, 2025.

\bibitem[Kong et~al.(2024)Kong, Hombaiah, Zhang, Mei, and Bendersky]{kong_prewrite_2024}
Weize Kong, Spurthi Hombaiah, Mingyang Zhang, Qiaozhu Mei, and Michael Bendersky.
\newblock {PRewrite}: {Prompt} {Rewriting} with {Reinforcement} {Learning}.
\newblock In \emph{Proceedings of the 62nd {Annual} {Meeting} of the {Association} for {Computational} {Linguistics} ({Volume} 2: {Short} {Papers})}, pages 594--601, Bangkok, Thailand, 2024. Association for Computational Linguistics.

\bibitem[Liu et~al.(2025)Liu, Chen, Li, Qi, Pang, Du, Lee, and Lin]{liu2025understandingr1zeroliketrainingcritical}
Zichen Liu, Changyu Chen, Wenjun Li, Penghui Qi, Tianyu Pang, Chao Du, Wee~Sun Lee, and Min Lin.
\newblock Understanding r1-zero-like training: A critical perspective, 2025.

\bibitem[Loshchilov and Hutter(2019)]{loshchilovdecoupled}
Ilya Loshchilov and Frank Hutter.
\newblock Decoupled weight decay regularization.
\newblock In \emph{International Conference on Learning Representations}, 2019.

\bibitem[Lyu et~al.(2024)Lyu, Shridhar, Malaviya, Zhang, Elazar, Tandon, Apidianaki, Sachan, and Callison-Burch]{lyu2024calibrating}
Qing Lyu, Kumar Shridhar, Chaitanya Malaviya, Li Zhang, Yanai Elazar, Niket Tandon, Marianna Apidianaki, Mrinmaya Sachan, and Chris Callison-Burch.
\newblock Calibrating large language models with sample consistency.
\newblock In \emph{AAAI Conference on Artificial Intelligence}, 2024.

\bibitem[Moor et~al.(2023)Moor, Banerjee, Abad, Krumholz, Leskovec, Topol, and Rajpurkar]{Moor2023FoundationMF}
Michael Moor, Oishi Banerjee, Zahra F~H Abad, Harlan~M. Krumholz, Jure Leskovec, Eric~J. Topol, and Pranav Rajpurkar.
\newblock Foundation models for generalist medical artificial intelligence.
\newblock \emph{Nature}, 616:\penalty0 259--265, 2023.

\bibitem[Nair et~al.(2020)Nair, Precup, Arnold, and Arbel]{nair2020exploring}
Tanya Nair, Doina Precup, Douglas~L Arnold, and Tal Arbel.
\newblock Exploring uncertainty measures in deep networks for multiple sclerosis lesion detection and segmentation.
\newblock \emph{Medical image analysis}, 59:\penalty0 101557, 2020.

\bibitem[Ouyang et~al.(2022)Ouyang, Wu, Jiang, Almeida, Wainwright, Mishkin, Zhang, Agarwal, Slama, Ray, et~al.]{ouyang2022training}
Long Ouyang, Jeffrey Wu, Xu Jiang, Diogo Almeida, Carroll Wainwright, Pamela Mishkin, Chong Zhang, Sandhini Agarwal, Katarina Slama, Alex Ray, et~al.
\newblock Training language models to follow instructions with human feedback.
\newblock \emph{Advances in Neural Information Processing Systems}, 35:\penalty0 27730--27744, 2022.

\bibitem[Pakdaman~Naeini et~al.(2015)Pakdaman~Naeini, Cooper, and Hauskrecht]{pakdaman_naeini_obtaining_2015}
Mahdi Pakdaman~Naeini, Gregory Cooper, and Milos Hauskrecht.
\newblock Obtaining {Well} {Calibrated} {Probabilities} {Using} {Bayesian} {Binning}.
\newblock \emph{Proceedings of the AAAI Conference on Artificial Intelligence}, 29\penalty0 (1), 2015.

\bibitem[Qwen et~al.(2025)Qwen, :, Yang, Yang, Zhang, Hui, Zheng, Yu, Li, Liu, Huang, Wei, Lin, Yang, Tu, Zhang, Yang, Yang, Zhou, Lin, Dang, Lu, Bao, Yang, Yu, Li, Xue, Zhang, Zhu, Men, Lin, Li, Tang, Xia, Ren, Ren, Fan, Su, Zhang, Wan, Liu, Cui, Zhang, and Qiu]{qwen2025qwen25technicalreport}
Qwen, :, An Yang, Baosong Yang, Beichen Zhang, Binyuan Hui, Bo Zheng, Bowen Yu, Chengyuan Li, Dayiheng Liu, Fei Huang, Haoran Wei, Huan Lin, Jian Yang, Jianhong Tu, Jianwei Zhang, Jianxin Yang, Jiaxi Yang, Jingren Zhou, Junyang Lin, Kai Dang, Keming Lu, Keqin Bao, Kexin Yang, Le Yu, Mei Li, Mingfeng Xue, Pei Zhang, Qin Zhu, Rui Men, Runji Lin, Tianhao Li, Tianyi Tang, Tingyu Xia, Xingzhang Ren, Xuancheng Ren, Yang Fan, Yang Su, Yichang Zhang, Yu Wan, Yuqiong Liu, Zeyu Cui, Zhenru Zhang, and Zihan Qiu.
\newblock Qwen2.5 technical report, 2025.

\bibitem[Shao et~al.(2024)Shao, Wang, Zhu, Xu, Song, Bi, Zhang, Zhang, Li, Wu, et~al.]{shao2024deepseekmath}
Zhihong Shao, Peiyi Wang, Qihao Zhu, Runxin Xu, Junxiao Song, Xiao Bi, Haowei Zhang, Mingchuan Zhang, YK Li, Y Wu, et~al.
\newblock Deepseekmath: Pushing the limits of mathematical reasoning in open language models.
\newblock \emph{arXiv preprint arXiv:2402.03300}, 2024.

\bibitem[Shen et~al.(2025)Shen, Huang, Nichyporuk, and Arbel]{shen2025improving}
Xing Shen, Hengguan Huang, Brennan Nichyporuk, and Tal Arbel.
\newblock Improving robustness and reliability in medical image classification with latent-guided diffusion and nested-ensembles.
\newblock \emph{IEEE Transactions on Medical Imaging}, 2025.

\bibitem[Singhal et~al.(2022)Singhal, Azizi, Tu, Mahdavi, Wei, Chung, Scales, Tanwani, Cole-Lewis, Pfohl, Payne, Seneviratne, Gamble, Kelly, Scharli, Chowdhery, Mansfield, y~Arcas, Webster, Corrado, Matias, Chou, Gottweis, Toma{\vs}ev, Liu, Rajkomar, Barral, Semturs, Karthikesalingam, and Natarajan]{Singhal2022LargeLM}
K. Singhal, Shekoofeh Azizi, Tao Tu, Said Mahdavi, Jason Wei, Hyung~Won Chung, Nathan Scales, Ajay~Kumar Tanwani, Heather~J. Cole-Lewis, Stephen~J. Pfohl, P~A Payne, Martin~G. Seneviratne, Paul Gamble, Chris Kelly, Nathaneal Scharli, Aakanksha Chowdhery, P.~A. Mansfield, Blaise~Ag{\"u}era y Arcas, Dale~R. Webster, Greg~S. Corrado, Yossi Matias, Katherine Hui-Ling Chou, Juraj Gottweis, Nenad Toma{\vs}ev, Yun Liu, Alvin Rajkomar, Jo{\"e}lle~K. Barral, Christopher Semturs, Alan Karthikesalingam, and Vivek Natarajan.
\newblock Large language models encode clinical knowledge.
\newblock \emph{Nature}, 620:\penalty0 172 -- 180, 2022.

\bibitem[Team(2024)]{qwen2.5}
Qwen Team.
\newblock Qwen2.5: A party of foundation models, 2024.

\bibitem[Team(2025)]{qwen2.5-VL}
Qwen Team.
\newblock Qwen2.5-vl, 2025.

\bibitem[Thirunavukarasu et~al.(2023)Thirunavukarasu, Ting, Elangovan, Gutierrez, Tan, and Ting]{Thirunavukarasu2023LargeLM}
Arun~James Thirunavukarasu, Darren Shu~Jeng Ting, Kabilan Elangovan, Laura Gutierrez, Ting~Fang Tan, and Daniel Shu~Wei Ting.
\newblock Large language models in medicine.
\newblock \emph{Nature Medicine}, 29:\penalty0 1930--1940, 2023.

\bibitem[Tian et~al.(2023)Tian, Mitchell, Zhou, Sharma, Rafailov, Yao, Finn, and Manning]{tian2023just}
Katherine Tian, Eric Mitchell, Allan Zhou, Archit Sharma, Rafael Rafailov, Huaxiu Yao, Chelsea Finn, and Christopher~D Manning.
\newblock Just ask for calibration: Strategies for eliciting calibrated confidence scores from language models fine-tuned with human feedback.
\newblock In \emph{Proceedings of the 2023 Conference on Empirical Methods in Natural Language Processing}, pages 5433--5442, 2023.

\bibitem[Wang et~al.(2024{\natexlab{a}})Wang, Bai, Tan, Wang, Fan, Bai, Chen, Liu, Wang, Ge, Fan, Dang, Du, Ren, Men, Liu, Zhou, Zhou, and Lin]{wang2024qwen2vlenhancingvisionlanguagemodels}
Peng Wang, Shuai Bai, Sinan Tan, Shijie Wang, Zhihao Fan, Jinze Bai, Keqin Chen, Xuejing Liu, Jialin Wang, Wenbin Ge, Yang Fan, Kai Dang, Mengfei Du, Xuancheng Ren, Rui Men, Dayiheng Liu, Chang Zhou, Jingren Zhou, and Junyang Lin.
\newblock Qwen2-vl: Enhancing vision-language model's perception of the world at any resolution, 2024{\natexlab{a}}.

\bibitem[Wang et~al.(2024{\natexlab{b}})Wang, Chen, Wang, Cao, Liu, Gao, Zhu, Zhu, Lu, Qiao, and Dai]{wang2024mpo}
Weiyun Wang, Zhe Chen, Wenhai Wang, Yue Cao, Yangzhou Liu, Zhangwei Gao, Jinguo Zhu, Xizhou Zhu, Lewei Lu, Yu Qiao, and Jifeng Dai.
\newblock Enhancing the reasoning ability of multimodal large language models via mixed preference optimization.
\newblock \emph{arXiv preprint arXiv:2411.10442}, 2024{\natexlab{b}}.

\bibitem[Wu et~al.(2024{\natexlab{a}})Wu, Lin, Zhang, Zhang, Xie, and Wang]{wu2024pmc}
Chaoyi Wu, Weixiong Lin, Xiaoman Zhang, Ya Zhang, Weidi Xie, and Yanfeng Wang.
\newblock Pmc-llama: toward building open-source language models for medicine.
\newblock \emph{Journal of the American Medical Informatics Association}, 31\penalty0 (9):\penalty0 1833--1843, 2024{\natexlab{a}}.

\bibitem[Wu et~al.(2024{\natexlab{b}})Wu, Lin, Dai, Hu, Shu, Ng, Jaillet, and Low]{wu2024prompt}
Zhaoxuan Wu, Xiaoqiang Lin, Zhongxiang Dai, Wenyang Hu, Yao Shu, See-Kiong Ng, Patrick Jaillet, and Bryan Kian~Hsiang Low.
\newblock Prompt optimization with {EASE}? efficient ordering-aware automated selection of exemplars.
\newblock In \emph{The Thirty-eighth Annual Conference on Neural Information Processing Systems}, 2024{\natexlab{b}}.

\bibitem[Xiong et~al.(2024)Xiong, Hu, Lu, LI, Fu, He, and Hooi]{xiong2024can}
Miao Xiong, Zhiyuan Hu, Xinyang Lu, YIFEI LI, Jie Fu, Junxian He, and Bryan Hooi.
\newblock Can {LLM}s express their uncertainty? an empirical evaluation of confidence elicitation in {LLM}s.
\newblock In \emph{The Twelfth International Conference on Learning Representations}, 2024.

\bibitem[Yang et~al.(2024)Yang, Yang, Hui, Zheng, Yu, Zhou, Li, Li, Liu, Huang, Dong, Wei, Lin, Tang, Wang, Yang, Tu, Zhang, Ma, Xu, Zhou, Bai, He, Lin, Dang, Lu, Chen, Yang, Li, Xue, Ni, Zhang, Wang, Peng, Men, Gao, Lin, Wang, Bai, Tan, Zhu, Li, Liu, Ge, Deng, Zhou, Ren, Zhang, Wei, Ren, Fan, Yao, Zhang, Wan, Chu, Liu, Cui, Zhang, and Fan]{qwen2}
An Yang, Baosong Yang, Binyuan Hui, Bo Zheng, Bowen Yu, Chang Zhou, Chengpeng Li, Chengyuan Li, Dayiheng Liu, Fei Huang, Guanting Dong, Haoran Wei, Huan Lin, Jialong Tang, Jialin Wang, Jian Yang, Jianhong Tu, Jianwei Zhang, Jianxin Ma, Jin Xu, Jingren Zhou, Jinze Bai, Jinzheng He, Junyang Lin, Kai Dang, Keming Lu, Keqin Chen, Kexin Yang, Mei Li, Mingfeng Xue, Na Ni, Pei Zhang, Peng Wang, Ru Peng, Rui Men, Ruize Gao, Runji Lin, Shijie Wang, Shuai Bai, Sinan Tan, Tianhang Zhu, Tianhao Li, Tianyu Liu, Wenbin Ge, Xiaodong Deng, Xiaohuan Zhou, Xingzhang Ren, Xinyu Zhang, Xipin Wei, Xuancheng Ren, Yang Fan, Yang Yao, Yichang Zhang, Yu Wan, Yunfei Chu, Yuqiong Liu, Zeyu Cui, Zhenru Zhang, and Zhihao Fan.
\newblock Qwen2 technical report.
\newblock \emph{arXiv preprint arXiv:2407.10671}, 2024.

\bibitem[Zhang et~al.(2024{\natexlab{a}})Zhang, Wu, Zhao, Lin, Zhang, Wang, and Xie]{zhang_development_2024}
Xiaoman Zhang, Chaoyi Wu, Ziheng Zhao, Weixiong Lin, Ya Zhang, Yanfeng Wang, and Weidi Xie.
\newblock Development of a large-scale medical visual question-answering dataset.
\newblock \emph{Communications Medicine}, 4\penalty0 (1):\penalty0 1--13, 2024{\natexlab{a}}.
\newblock Publisher: Nature Publishing Group.

\bibitem[Zhang et~al.(2024{\natexlab{b}})Zhang, Wu, Zhao, Lin, Zhang, Wang, and Xie]{zhang_pmc-vqa_2024}
Xiaoman Zhang, Chaoyi Wu, Ziheng Zhao, Weixiong Lin, Ya Zhang, Yanfeng Wang, and Weidi Xie.
\newblock {PMC}-{VQA}: {Visual} {Instruction} {Tuning} for {Medical} {Visual} {Question} {Answering}, 2024{\natexlab{b}}.
\newblock arXiv:2305.10415 [cs].

\bibitem[Zhao et~al.(2024)Zhao, Zhang, Pan, Yao, Yu, Wu, and Chen]{zhao2024fact}
Xinran Zhao, Hongming Zhang, Xiaoman Pan, Wenlin Yao, Dong Yu, Tongshuang Wu, and Jianshu Chen.
\newblock Fact-and-reflection (far) improves confidence calibration of large language models.
\newblock In \emph{Findings of the Association for Computational Linguistics ACL 2024}, pages 8702--8718, 2024.

\end{thebibliography}
}

\end{document}